\pdfoutput=1 % arxiv
\documentclass{article} % For LaTeX2e

\usepackage{iclr2020_conference,times}

\usepackage{amsfonts}
\usepackage{amssymb}
\usepackage{amsmath}
\usepackage{geometry}
\usepackage{graphicx}
\usepackage{physics} 
\usepackage{wrapfig}
\usepackage[protrusion=true,tracking=true,expansion,final]{microtype}      % microtypography
\usepackage{xcolor} %colors
\usepackage{booktabs} % for professional tables
\usepackage{subcaption}
\usepackage{nicefrac} % nice flat fraction
\usepackage{cancel}

\usepackage{gensymb}
\usepackage{bbding}
\usepackage{bm}
\usepackage{mathtools}
\usepackage{centernot} % not parallel, etc.
\usepackage{lmodern}
\usepackage{sectsty} % colorful sections
\usepackage{bigints}
\usepackage{dsfont} %mathbb 1
\usepackage{esint} % beatiful integrals
\usepackage{amsthm} % theorems
\usepackage{natbib}

\usepackage{multirow}
\usepackage{varioref} % references to show page
\usepackage[pagebackref]{hyperref}
\usepackage[capitalise]{cleveref} 

\crefname{appsec}{appendix}{appendices}
\Crefname{appsec}{Appendix}{Appendices}

\usepackage{url}

\renewcommand{\cite}[1]{\citep{#1}}
\definecolor{mydarkblue}{rgb}{0,0.08,0.45}
\definecolor{urlcolor}{rgb}{0,.145,.698}
\definecolor{linkcolor}{rgb}{.71,0.21,0.01}
\hypersetup{ %
	pdftitle={CAT: Compression-Aware Training for bandwidth reduction}, %TODO
	pdfauthor={Chaim Baskin, Brian Chmiel, Evgenii Zheltonozhskii, Ron Banner, Alex M. Bronstein, Avi Mendelson},
	pdfsubject={},
	pdfkeywords={},
	pdfborder=0 0 0,
	pdfpagemode=UseNone,
	breaklinks=true,  % so long urls are correctly broken across lines
	colorlinks=true,
	urlcolor=urlcolor,
	linkcolor=linkcolor,
	citecolor=mydarkblue,
	filecolor=mydarkblue,
	pdfview=FitH}

\renewcommand*{\backref}[1]{} % for backref < 1.33 necessary
\renewcommand*{\backrefalt}[4]{%
	\ifcase #1 %
	\or
	(cited on p. #2)%
	\else
	(cited on pp. #2)%
	\fi
}

\vbadness=\maxdimen
\usepackage{chngcntr}
\usepackage{apptools}
\AtAppendix{\counterwithin{figure}{section}}
\AtAppendix{\counterwithin{table}{section}}

\DeclarePairedDelimiter{\ceil}{\lceil}{\rceil}

\title{CAT: Compression-Aware Training for bandwidth reduction}

\newcommand*\samethanks[1][\value{footnote}]{\footnotemark[#1]}
\author{
Chaim Baskin\,${^\dagger}{^\circ}$\thanks{Equal contribution.}\quad 
Brian Chmiel\,${^\dagger}{^\circ}$\samethanks[1]\quad 
Evgenii Zheltonozhskii\,$^\circ$\samethanks[1]\quad
Ron Banner\,$^\dagger$
\\[0.15cm]\textbf{
Alex M. Bronstein\,$^\circ$\quad 
Avi Mendelson\,$^\circ$} 
\\[0.2cm]
$^\dagger$Intel   --   Artificial   Intelligence   Products   Group (AIPG), Haifa, Israel,\\ 
$^\circ$Technion -- Israel Institute of Technology, Haifa, Israel 
\\[0.2cm]
\small{\texttt{\{\href{mailto:brian.chmiel@intel.com}{brian.chmiel}, \href{mailto:ron.banner@intel.com}{ron.banner}\}@intel.com}}\\
\small{\texttt{\{\href{mailto:chaimbaskin@cs.technion.ac.il}{chaimbaskin},  \href{mailto:bron@cs.technion.ac.il}{bron}, \href{mailto:avi.mendelson@cs.technion.ac.il}{avi.mendelson}\}@cs.technion.ac.il}}\\
\small{\texttt{\{\href{mailto:evgeniizh@campus.technion.ac.il}{evgeniizh}\}@campus.technion.ac.il }}
}

\iclrfinalcopy 
\begin{document}

\maketitle

\begin{abstract}
Convolutional neural networks (CNNs) have become the dominant neural network architecture for solving visual processing tasks. One of the major obstacles hindering the ubiquitous use of CNNs for inference is their relatively high memory bandwidth requirements, which can be a main energy consumer and throughput bottleneck in hardware accelerators. Accordingly, an efficient feature map compression method can result in substantial performance gains. Inspired by \textit{quantization-aware training} approaches, we propose a compression-aware training (CAT) method that involves training the model in a way that allows better compression of feature maps during inference. Our method trains the model to achieve low-entropy feature maps, which enables efficient compression at inference time using classical transform coding methods. 
CAT significantly improves the state-of-the-art results reported for quantization. For example, on ResNet-34 we achieve 73.1\% accuracy (0.2\% degradation from the baseline)  with an average representation of only 1.79 bits per value.
\href{https://github.com/CAT-teams/CAT}{Reference implementation} accompanies the paper.
\end{abstract}

\section{Introduction}
\label{sec:intro}
Deep Neural Networks (DNNs) have become  a popular choice for a wide range of applications such as computer vision, natural language processing, autonomous cars, etc. Unfortunately, their vast demands for computational resources often prevents their use on power-challenged platforms. The desire for reduced bandwidth and compute requirements of deep learning models has driven research into quantization 
\cite{hubara2016quantized,yang2019quantization,liu2019rbcn, gong2019differentiable}, pruning
\cite{lecun1990optimal,li2016pruning,molchanov2019importance}, and sparsification \cite{gale2019state,dettmers2019sparse}.

In particular, quantization works usually focus on scalar quantization of the feature maps: mapping the activation values to a discrete set $\qty{q_i}$ of size $L$. Such a representation, while being less precise, is especially useful in custom hardware, where it allows more efficient computations and reduces the memory bandwidth. In this work, we focus on the latter, which has been shown to dominate the energy footprint of CNN inference on custom hardware \cite{energy}. We show that the quantized activation values $\qty{q_i}$ can further be coded to reduce memory requirements.

The raw quantized data require $\ceil{\log_2(L)}$ bits per value for storage. This number can be reduced by compressing the feature maps. In particular, in the case of element-wise compression of independent identically distributed values, the lower bound of amount of bits per element is given by the entropy \cite{shannon1948mathematical}:
\begin{align}
    H(\vb{q}) &= -\sum_{i=1}^{L} p(q_i)\log_2 p(q_i)
    \label{entropy}
\end{align}
of the quantized values $\qty{q_i}$, where $p(q_i)$ denotes the probability of $q_i$. 

\begin{figure}
 \centering
        \begin{subfigure}[c]{0.48\linewidth}
            \includegraphics[width=\linewidth]{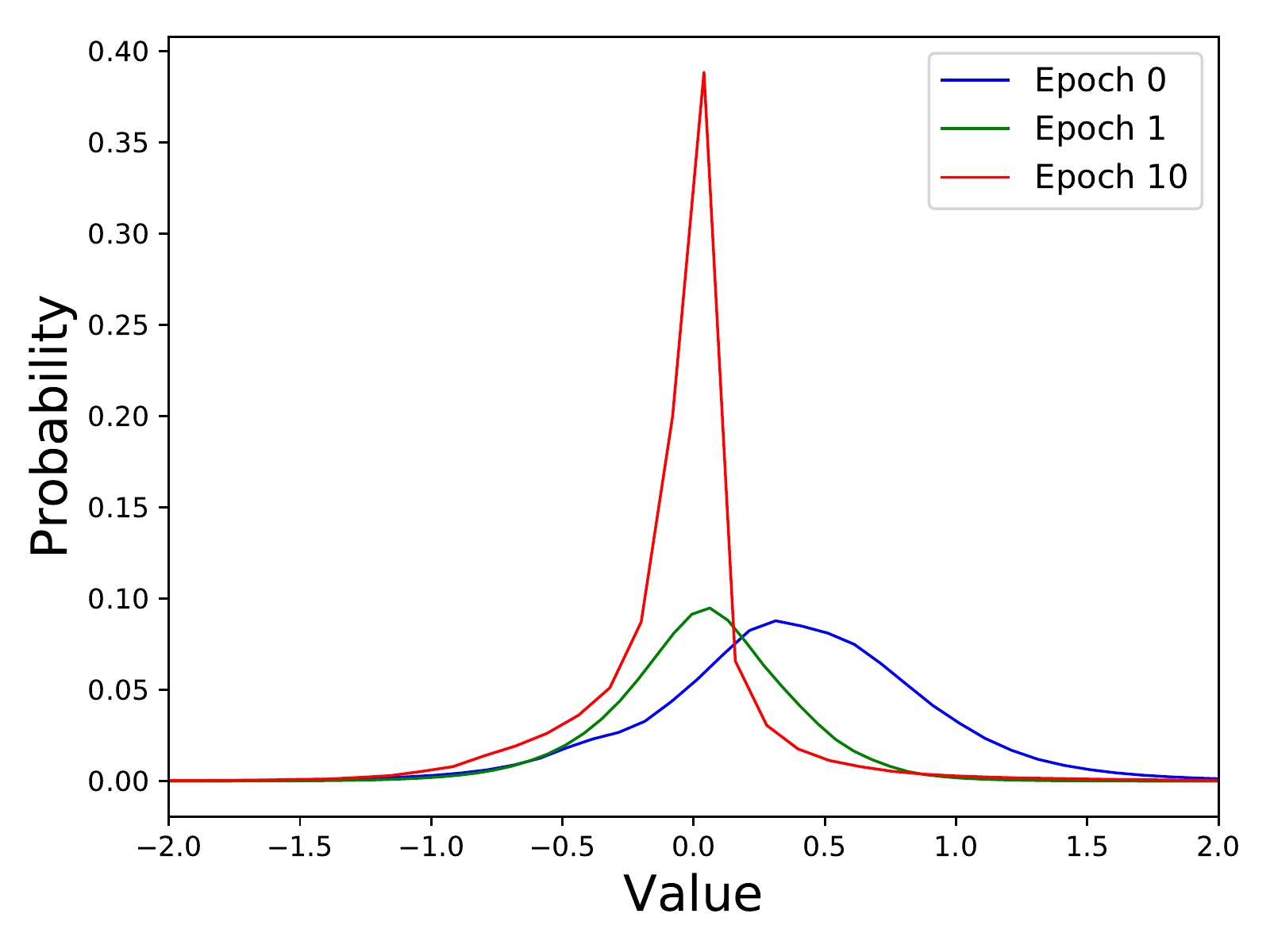}
            \subcaption{}
            \label{fig:act_hist_a}
        \end{subfigure} 
        \hfill
        \begin{subfigure}[c]{0.48\linewidth}
            \includegraphics[width=\linewidth]{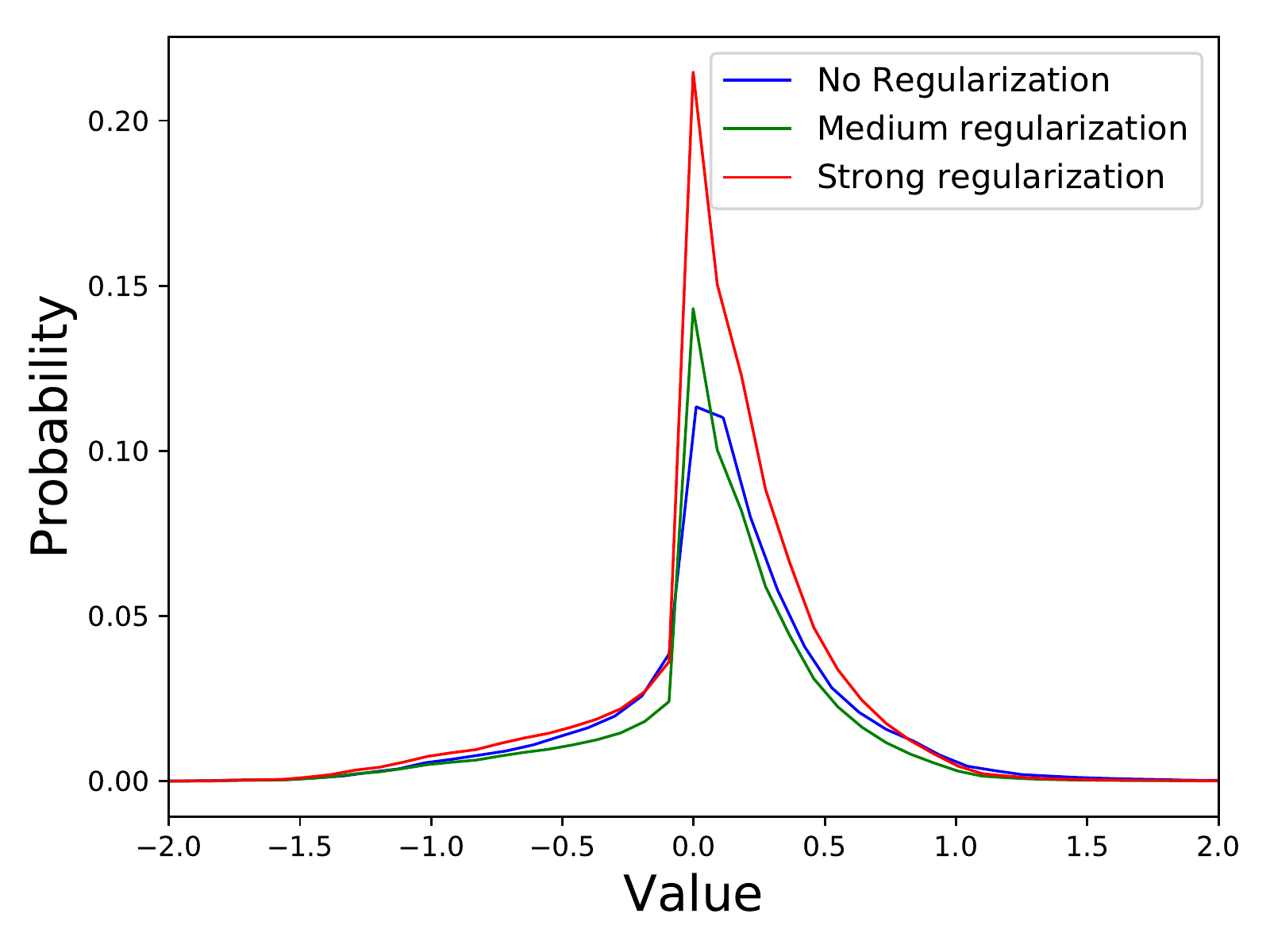}
            \subcaption{}
            \label{fig:act_hist_b}
        \end{subfigure} 
        \caption{ Pre-activation distributions of one layer in ResNet-18. \textbf{(a) Evolution at different epochs.} As training progresses, the probability of non-positive pre-activation values increases, zeroing more post-ReLU values. The sharp peak at zero reduces entropy and thus improves compressibility.
        \textbf{(b) Effect of entropy regularization.} Without regularization, the distribution has much heavier tails and thus has higher entropy.
        With increasing regularization, the probability of extreme values is significantly reduced. The entropy penalty $\lambda$ was selected so that the overall accuracy is not affected. The compression ratio in the strongly regularized case is $2.23$ higher compared to the unregularized the baseline. 
        }  
        \label{fig:act_hist}
\end{figure}

In this work, we take a further step by manipulating the distribution of the quantization values so that the entropy $H(q)$ is minimized. To that end, we formulate the training problem by augmenting the regular task-specific loss (the cross-entropy classifier loss in our case) with the feature map entropy serving as a proxy to the memory rate. The strength of the latter penalty is controlled through a parameter $\lambda>0$. \cref{fig:act_hist} demonstrates the effect of the entropy penalty on the compressibility of the intermediate activations. 

\paragraph{Contributions.} Our paper makes several contributions. 
Firstly, we introduce Compression-Aware Training (CAT), a technique for memory bandwidth reduction. The method works by introducing a loss term that penalizes the entropy of the activations at training time and applying entropy encoding (e.g., Huffman coding) on the resulting activations at inference time. Since the only overhead of the method at inference time is entropy encoding that can be implemented efficiently, the improvement is universal for any hardware implementation, being especially efficient on computationally optimized ones, where memory I/O dominates the energy footprint \cite{energy, jouppi2017datacenter}. We demonstrate a $2$ to $4$-fold memory bandwidth reduction for multiple architectures: MobileNetV2 and ResNet on the ImageNet visual recongition task, and SSD512 on the PASCAL VOC object detection task.
We also investigate several differentiable loss functions which lead to activation entropy minimization and show a few alternatives that lead to the same effect. 

Finally, we analyze the rate-distortion tradeoff of the method, achieving even stronger compression at the expense of minor accuracy reduction: for ResNet-18, we manage to achieve entropy inferior to one bit per value, at the expense of losing 2\% of the top-1 accuracy.

%  \paragraph{Contributions.} In this paper we propose a way to reduce a memory bandwidth by introducing a loss term that penalizes the entropy of the activations in training time and applying entropy encoding on them in inference time. We show that different forms of such loss lead to same effect and show \ez{more than $3 \times$} improvement for multiple architectures: MobileNet v2 and ResNet on ImageNet and SSD512 on PASCAL VOC dataset. In addition, we analyze rate-distortion tradeoff of the method, achieving even stronger compression at the expense of minor accuracy reduction. Since the only overhead of the method in the inference time is entropy encoding, which can be implemented efficiently, the improvement is universal for any hardware implementation, being especially efficient on computationally optimized ones, where memory I/O is a main energy consumer \cite{energy, jouppi2017datacenter}. 
\section{Related work}
\label{sec:related_work}

Recent studies \cite{energy, wang2019lutnet} have shown that almost 70\% of the energy footprint on custom hardware is due to the data movement to and from the off-chip memory.
Nonetheless, the techniques for memory bandwidth reduction have not yet received significant attention in the literature. 
One way to improve memory performance is by fusing convolutional layers \citet{Xiao:2017:EHA:3061639.3062244}, \citet{xing2019dnnvm}, reducing the number of feature maps transfers. This reduces both runtime and energy consumption of the hardware accelerator. Another option is to use on-chip cache instead of external memory. \citet{Morcel:2019:FAC:3322884.3306202} have shown order of magnitude improvement in power consuption using this technique.
Another important system parameter dominated by the memory bandwidth is latency. \citet{jouppi2017datacenter} and \citet{wang2019lutnet} showed that the state-of-the-art  DNN accelerators are memory-bound, implying that increasing computation throughput without reducing the memory bandwidth barely affects the total system latency.

 Quantization reduces computation and memory requirements; 16-bit fixed point has become a \emph{de facto} standard for fast inference. In most applications, weights and activations can be quantized down to $8$ bits without noticeable loss of precision \cite{lee2018quantization, yang2019swalp}. Further quantization to lower precision requires non-trivial techniques \cite{mishra2017wrpn,ZhangYangYeECCV2018}, which are currently capable of reaching around $3-4$ without compromising precision
 \cite{choi2018pact, choi2018bridging, dong2019hawq}. 
 %  \cite{rastegari2016xnor,hubara_quant}
 % \cite{liu2019rbcn,peng2019bdnn}

%However,those works require fine-tuning or training from scratch which means you need both labeled dataset and time. 
%\citet{migacz20178}\citet{Gong_2018}\citet{lee2018quantization}\citet{banner2018posttraining}\citet{meller2019different}\citet{choukroun2019lowbit}

%[WHY ARE WE WRITING THIS? IT COMPLETELY UNDERMINES OUR GOALS]
%Energy consuption is an important characteristic of the custom hardware. \citet{energy} noted that 70\% of consumed energy is spent on memory transfers.
%\citet{ansari2018selective}, on the other hand, have seen almost no power reduction after optimizing memory bandwidth. It suggests that due to the fact that usually universal memory modules are used, the optimization of the computational part plays main role in energy consumption distribution.

Another way to reduce memory bandwidth is by compressing the intermediate activations prior to their transfer to memory with some computationally cheap encoding, such that Huffman \cite{chandra2018data, chmiel2019feature} or run-length (RLE) encoding \cite{cavigelli2019ebpc}. 
%In particular, Huffman coding is known to be nearly optimal for element-by-element coding. 
A similar approach of storing only nonzero values was utilized by \citet{lin2018supporting}.
\cite{chmiel2019feature} used linear dimensionality reduction (PCA) to increase the effectiveness of Huffman coding, while \citet{Gudovskiy_2018_ECCV_Workshops}  proposed to use nonlinear dimensionality reduction techniques.
 
Lossless coding was previously utilized in a number of ways for DNN compression: \citet{han2015deep} and \citet{zhao2019efficient} used Huffman coding to compress weights, while \citet{wijayanto2019towards} 
used more complicated DEFLATE (LZ77 + Huffman) algorithm for the same purpose. \citet{aytekin2019compressibility} proposed to use compressibility loss, which induces sparsity and has been shown (empirically) to reduce entropy of the non-zero part of the activations.

%\citet{agustsson2017soft} proposed a differentiable approximation of entropy by using soft quantization.

\section {Method}
\label{sec:method}
%\subsection{Problem Formulation}
We consider a feed-forward DNN $\mathcal{F}$ composed of $L$ layers; each subsequent layer processing the output of the previous one: $x^i =\mathcal{F}_i\qty(x^{i-1}) $, using the parameters $\vb{w} \in \mathbb{R}^N$. We denote by $x^0 = x$  and  $x^L = y$ the input and the output of the network, respectively.

The parameters  $\vb{w}$ of the network are learned by  minimizing  $\mathcal{L}(x,y;\vb{w}) + \lambda \mathcal{R}(\vb{w})$, with the former term $\mathcal{L}$ being the task loss, and the latter term $\mathcal{R}$ being a regularizer (e.g., $\|\vb{w} \|_2$) inducing some properties on the parameters $\vb{w}$.

\subsection{Entropy encoding and rate regularization}
Entropy encoders are a family of lossless data encoders which compress each symbol independently. In this case, assuming i.i.d. distribution of the input, it has been show that optimal code length is $-\log_b p$, where $b$ is number of symbols and $p_i$ is the probability of i$^{\text{th}}$ symbol \cite{shannon1948mathematical}. Thus, for a discrete random variable $X$ we define an entropy $H(X) = - \mathbb{E} \log_2 X = - \sum_{i} p(x_i) \log_2 p(x_i)$, which is a lower bound of the amount of information required for lossless compression of  $X$. 
The expected total space required to encode the message is $N \cdot H$, where $N$ is number of symbols.  Since we encode the activations with entropy encoder before writing them into memory, we would like to minimize the entropy of the activations  to improve the compression rate.

One example of entropy encoder is Huffman coding -- a prefix coding which assigns shorter codes to the more probable symbols. It is popular because of simplicity along with high compression rate \cite{szpankowski2000huffman} bounded by $H(X) \le R \le H(X)+1$  (the comparison of Huffman coding rates to entropy is shown in \cref{fig:acc_entropy}). 
Another entropy encoder is arithmetic coding -- highly efficient encoder which achieves optimal rates for big enough input but requires more computational resources for encoding.

\subsection{Differentiable entropy-reducing loss}
Since the empirical entropy is a discrete function, it is not differentiable and thus cannot be directly minimized with gradient descent. Nevertheless, there exist a number of differentiable functions which either approximate entropy or have same minimizer. Thus we optimize 
\begin{align}
    \mathcal{L} = \mathcal{L}_{\text{p}} + \lambda \mathcal{L}_{H},
    \label{eq:losses}
\end{align}
where $\mathcal{L}_{\text{p}}$ is a target loss function and $\mathcal{L}_{H}$ is some regularization which minimizes entropy.

%\subsubsection{Soft entropy}
\paragraph{Soft entropy} First, we consider the differentiable entropy estimation suggested by \citet{agustsson2017soft}. Ws start from definition of the entropy,
\begin{align}
    H(X) &= - \sum p(x_i) \log(p(x_i))\\
    p_i &= \frac{\abs{\qty{x|x=q_i}}}{N},
\end{align}
where $\vb{q}$ is a vector of quantized values.
%, i.e., $q_i$ is the center of bin $i$ in the quantization. 
Let $m$ be an index of the bin the current value is mapped to, and $\vb{Q}$ a one-hot encoding of this index, i.e.,
\begin{align}
q_m &= \arg\min\limits_{q_i \in \mathcal{Q}} \abs{x-q_i}  = \arg\max\limits_{q_i \in \mathcal{Q}} \qty(-\abs{x-q_i})\\
    \vb{Q}_i &= \delta_{im},
\end{align}
where $\delta_{im}$ denotes Kroneker's delta. To make the latter expression differentiable, we can replace argmax with softmax:
\begin{align}
    \tilde{\vb{Q}}(x) = \mathrm{softmax}\qty(-\abs{\vb{x}-\vb{q}}, T),
\end{align}
where $T$ is the temperature parameter, and $\tilde{Q}(x)\to Q(x)$ as  $T\to\infty$.

Finally, the soft entropy $\hat{H}$ is defined as
\begin{align}
    \hat{p}_i &= \frac{\sum_j \tilde{\vb{Q}}_{i}(x_j)}{N}\\
    \hat{H}(X) &= - \sum \hat{p}(x_i) \log(\hat{p}(x_i)) 
    \label{eq:diff_entropy}
\end{align}
To improve both memory requirements and time complexity of the training, we calculate soft entropy only on part of the batch, reducing the amount of computation and the gradient tensor size.

%\subsubsection{Compressibility loss for entropy reduction}
\paragraph{Compressibility loss for entropy reduction} An alternative loss promoting entropy reduction was proposed by \citet{aytekin2019compressibility} under the name of \emph{compressibility loss} and based on earlier work by \citet{hoyer2004non}:
\begin{align}
    \mathcal{L}_c = \frac{\norm{\vb{x}}_1}{\norm{\vb{x}}_2}
    \label{eq:comp_loss}
\end{align}
This loss has the advantage of computational simplicity, and has been shown both theoretically and practically to promote sparsity and low entropy in input vectors. While originally applied to the weights of the network, here we apply the same loss to the activations.

As shown in \cref{sec:ablation}, both the soft entropy and the compressibility loss lead to similar results.

%\subsection{Putting it all together}
Our method for reducing memory bandwidth can be described as follows: at training time, we fine-tune (training from scratch should also be possible, but we have not tested it) the pre-trained network $\mathcal{F}$ with the regularized loss (\ref{eq:losses}), where we use $\mathcal{L}_H = \sum \hat{H}(x_i)$ in case of differentiable entropy and $\mathcal{L}_H = \sum \mathcal{L}_c(x_i)$ in case of compressability loss. At test time, we apply entropy coding  on the activations before writing them to memory, thus reducing the amount of memory transactions. In contrast to \citet{chmiel2019feature}, who avoided fine-tuning by using test time transformation in order to reduce entropy, our method does not requires complex transformations at test time by inducing low entropy during training.
\section{Experimental Results}
\label{sec:expriments}

We evaluate the proposed scheme on common CNN architectures for image classification on ImageNet (ResNet-18/34/50, MobileNetV2), and object detection on Pascal VOC dataset (SSD512\footnote{Our code is based on implementation by \citet{lufficc2018ssd}.} \cite{liu2016ssd}). 
The weights were initialized with pre-trained model and quantized with uniform quantization using the shadow weights, i.e. applying updates to a full precision copy of quantized weights \cite{hubara2016quantized, rastegari2016xnor}. The activation were clipped with a learnable parameter and then uniform quantized as suggested by \citet{nice2018baskin}. Similarly to previous works \cite{dorefa2016,rastegari2016xnor}, we used the straight-through estimator \cite{bengio2013ste} to approximate the gradients.
We quantize all layers in the network, in contrast to the common practice of leaving first and last layer in high-precision \cite{dorefa2016,nice2018baskin}. Since the weights are quantized only to 8 bit, we have noticed small to no difference between the quantized and non-quantized weights. 

For optimization, we use SGD with a learning rate of $10^{-4}$, momentum $0.9$, and weight decay $4\times 10^{-5}$ for up to $30$ epochs (usually, $10-15$ epochs were sufficient for convergence). Our initial choice of temperature was $T=10$, which performed well. 
We tried, following the common approach \cite{jang2016categorical}, to apply exponential scheduling to the temperature, but it did not have any noticeable effect on the results.

\begin{figure}
 \centering
 
    \begin{subfigure}[b]{0.48\linewidth}
        \includegraphics[width=\linewidth]{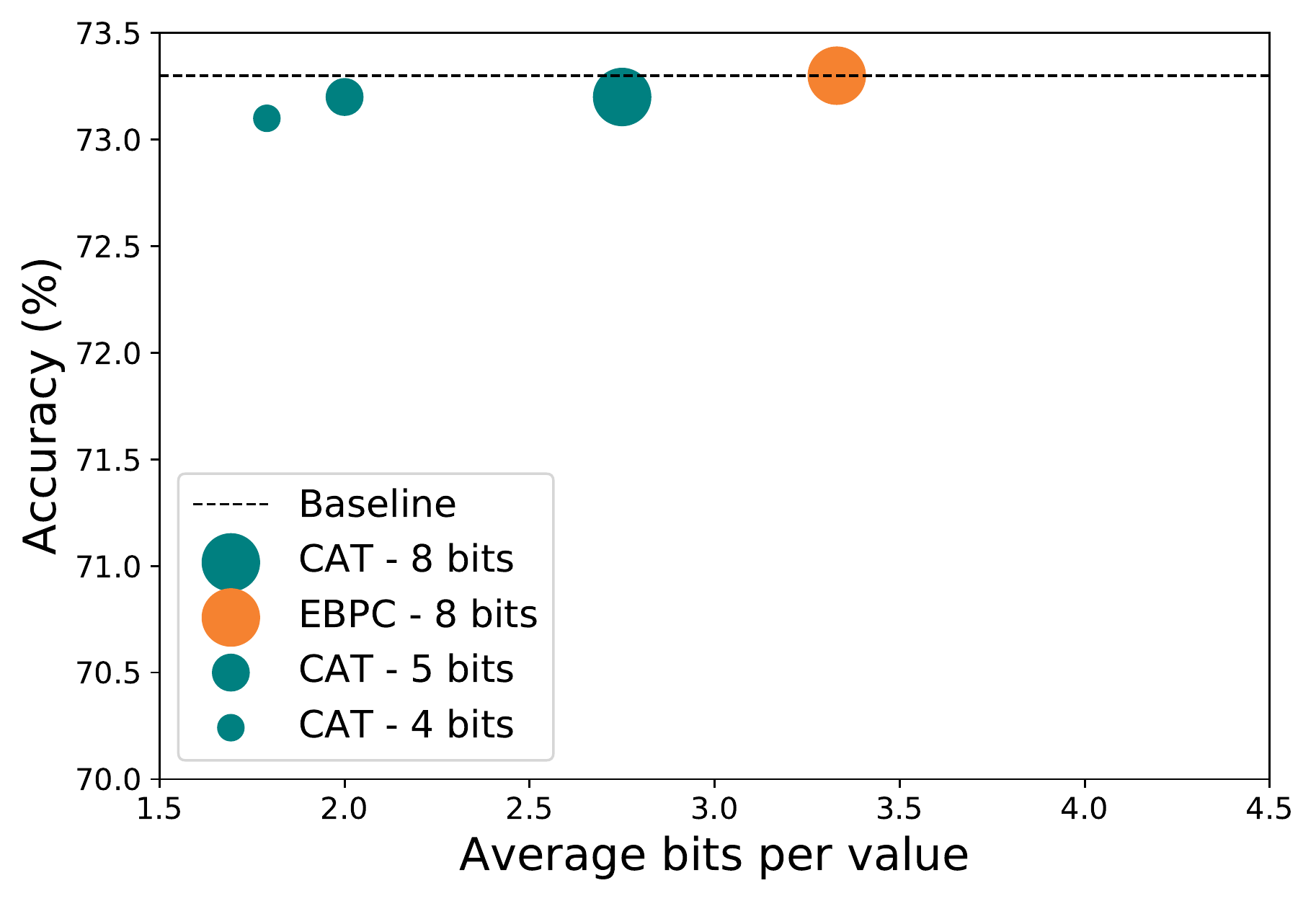}
        \subcaption{ }
        \label{resnet18-compare}
    \end{subfigure}
    \begin{subfigure}[b]{0.48\linewidth}
        \includegraphics[width=\linewidth]{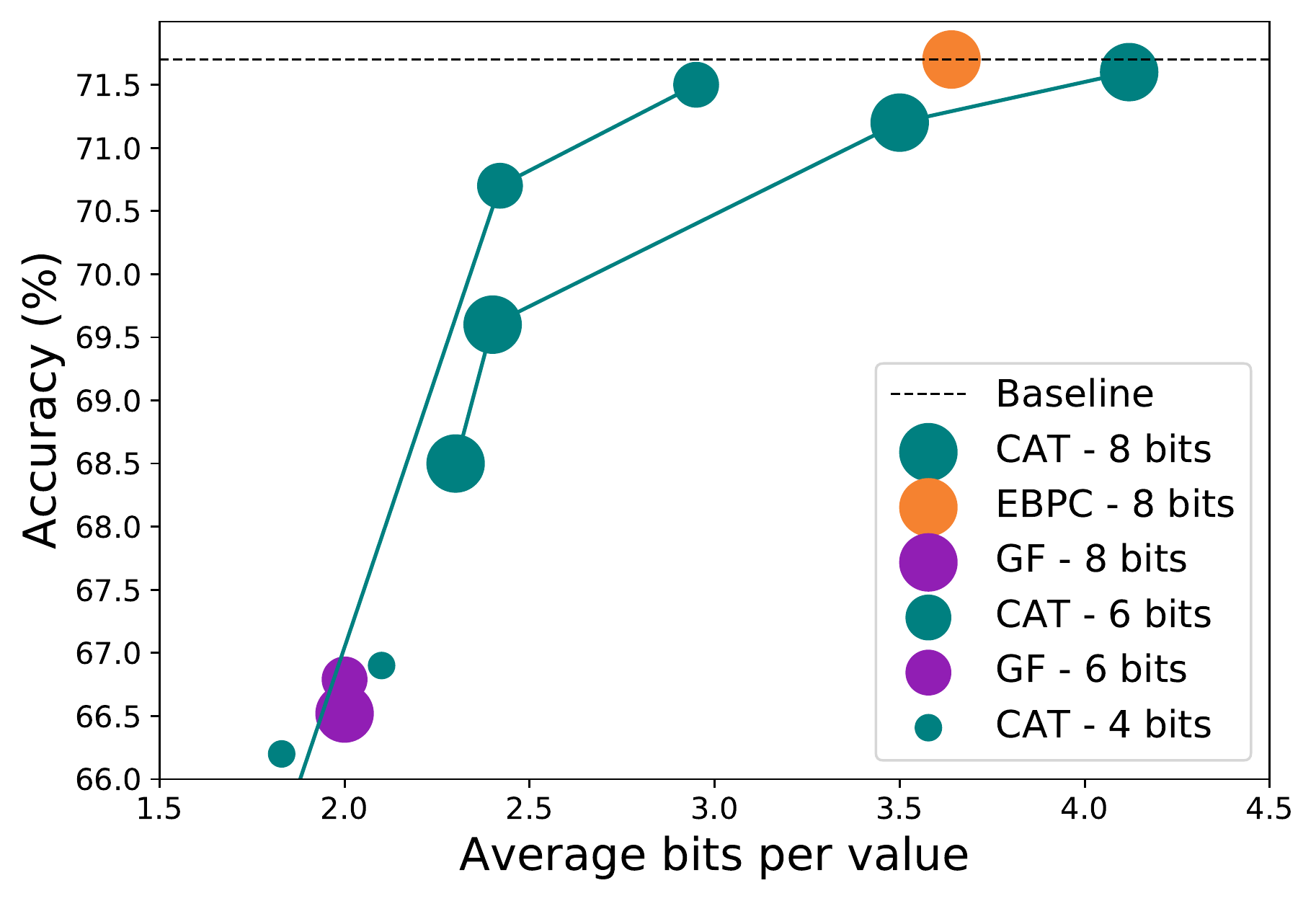}
        \subcaption{ }
        \label{mobilenet-compare}
    \end{subfigure}
 \caption{Comparison with other methods: EPBC (\citet{cavigelli2019ebpc}) and GF (\citet{Gudovskiy_2018_ECCV_Workshops}) in \textbf{(a) ResNet-34} and \textbf{(b) MobilenetV2}. Different marker size refers to different activation bitwidths before compression. For GF, compression rate was averaged only over compressed layers.
 %, which constitutes the compute overheads. 
 }
\label{fig:comparison}
\end{figure}

\begin{table}
    \centering
    \caption{
    %Comparison with other methods. \citet{cavigelli2019ebpc} can be used without fine-tuning. \citet{Gudovskiy_2018_ECCV_Workshops} compressed only part of the layers. For MobileNetV2, we provide results with two different levels of regularization. Weight bitwidth is 8 except for full-precision experiments. 
    Results %of our method 
    for ResNet-18, ResNet-50, and SSD512. For comparison we include the results obtained by \citet{Gudovskiy_2018_ECCV_Workshops} for the SSD512 model on the same task but with a different backbone, for which we obtain a better compression rate with a lower accuracy degradation. Compute denotes activation bitwidth %before compression 
    used for arithmetic operations. Memory denotes average number of bits for memory transactions (after compression). Compression ratio denotes the reduction in representation size. %(before and after compression).
    %Activation bitwidth denotes number of bits of activations used for computations, while average bits per value denotes amount of bits after compression. 
    Weight bitwidth is 8 except the full-precision experiments.
    Additional experimental results are provided in \cref{app:results}. }
    \begin{tabular}{lcccc}
        \toprule
        \textbf{Architecture} & \textbf{Compute}  & \textbf{Memory} & \textbf{Compression} & \textbf{Top-1 } \\
         & \textbf{(bits)}  & \textbf{(bits)} & \textbf{ratio} & \textbf{accuracy (\%)} \\
            \midrule
            
            \multirow{3}{*}{ResNet-18, CAT} &32&32& 1 & 69.70\\
             &5 &1.5& 3.33& 69.20\\
             &4 & 1.51 & 2.65 & 68.08\\
            \midrule
            
            % MobileNetV2, ours&32&32& 1 & 71.70 \\
            % \cmidrule{2-5}
            % MobileNetV2, ours&8& 4.12 &  1.94 &  71.60\\
            % MobileNetV2, ours&8& 2.40 &  3.33 &  69.60\\  
            % MobileNetV2 \cite{cavigelli2019ebpc}&8&3.64& 2.2 & \textbf{71.70}\\  
            % MobileNetV2 \cite{Gudovskiy_2018_ECCV_Workshops}&8&\textbf{2}& 4 & 66.6\\
            % \cmidrule{2-5}
            % MobileNetV2, ours&6& 2.95 & 2.03 & \textbf{71.50}\\
            % MobileNetV2, ours&6& 2.45 & 2.45 & 70.70 \\
            % MobileNetV2 \cite{Gudovskiy_2018_ECCV_Workshops}&6&6& 1 & 70.9\\
            % MobileNetV2 \cite{Gudovskiy_2018_ECCV_Workshops}&6&\textbf{2}& 3 & 66.7\\
            % \cmidrule{2-5}
            % MobileNetV2, ours&4& 2.10&  1.90 &  \textbf{66.90}\\
            % MobileNetV2, ours&4& \textbf{1.83} & 2.19  &  66.20 \\
            % \midrule
             
            % ResNet34, ours&32&32& 1 & 73.3\\
            % ResNet34, ours&8& 2.75 & 3.27  &73.2\\
            % ResNet34 \cite{cavigelli2019ebpc}&8&3.33& 2.4 & \textbf{73.3}\\
            % ResNet34, ours&5& 2.00 & 2.50 &73.2\\
            % ResNet34, ours&4& \textbf{1.79} & 2.23  &73.1\\
            % \midrule
            
            \multirow{3}{*}{ResNet-50, CAT}&32&32& 1 & 76.1 \\
            &5&  1.60& 3.125 & 74.90\\
            &4& 1.78 & 2.25  &74.50\\
            \midrule
            \multirow{3}{*}{\shortstack[l]{SSD512-SqueezeNet\\ \cite{Gudovskiy_2018_ECCV_Workshops}}}&32&32& 1 & 68.12\\
            &8&2& 4 & 64.39\\
            &6&2& 3 & 62.09\\
            
            \midrule
            \multirow{3}{*}{SSD512-VGG, CAT}&32&32& 1 & 80.72\\
            &6&2.334& 2.57 & 77.49 \\
            &4& 1.562 & 2.56 & 77.43 \\
         
            \bottomrule            
    \end{tabular}
    \label{tab:comparison}
\end{table}

In \cref{fig:comparison} we compare our method with EPBC (\citet{cavigelli2019ebpc}) and GF (\citet{Gudovskiy_2018_ECCV_Workshops}). 
 EPBC is based on a lossless compression method that maintains the full precision accuracy while reducing bit rate to approximately 3.5 bits/value in both models. GF, on the other hand, provides strong compression at the expense of larger accuracy degradations.  In addition, \citet{Gudovskiy_2018_ECCV_Workshops} have compressed only part of the layers.
 Unlike these two methods, CAT allows more flexible tradeoff between compression and accuracy. CAT shows better results in ResNet-34 and show either better accuracy or compression for MobileNetV2.
We also run our method on additional architectures: ResNet-18, ResNet-50, and SSD512 with VGG backbone; the results are listed in \cref{tab:comparison}. Even though we can not directly compare detection results with \citet{Gudovskiy_2018_ECCV_Workshops}, the drop in accuracy is lower in our case. 

\subsection{Ablation study}
\label{sec:ablation}
%\ez{\cref{fig:act_hist}}
%\ez{per-layer entropy after training. with and without reg?}

\begin{figure}
 \centering \vspace{-4mm}
    \begin{subfigure}[b]{0.48\linewidth}
        \includegraphics[width=\linewidth]{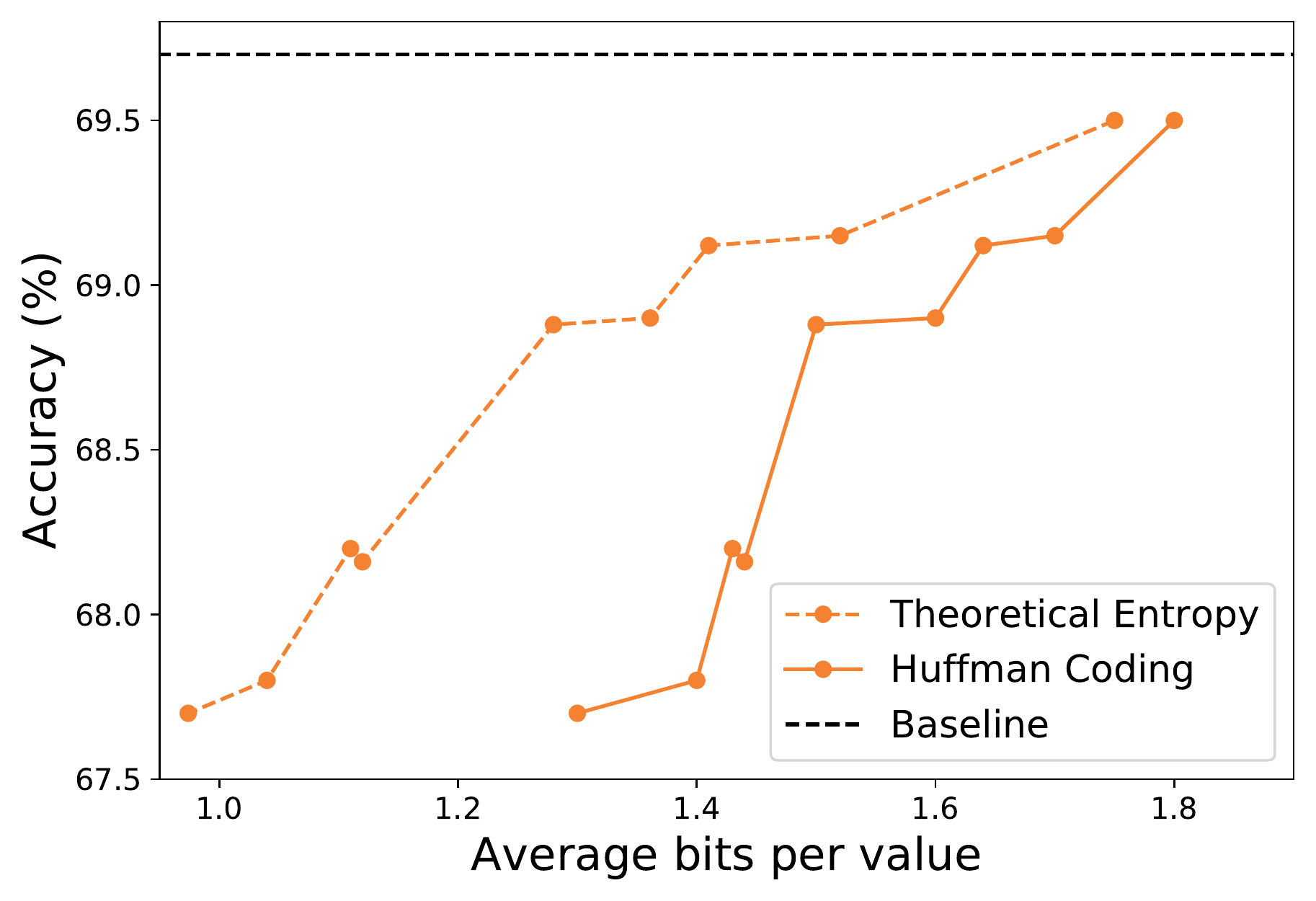}
        \subcaption{ }
        \label{resnet18-tradeoff}
    \end{subfigure}
    \begin{subfigure}[b]{0.48\linewidth}
        \includegraphics[width=\linewidth]{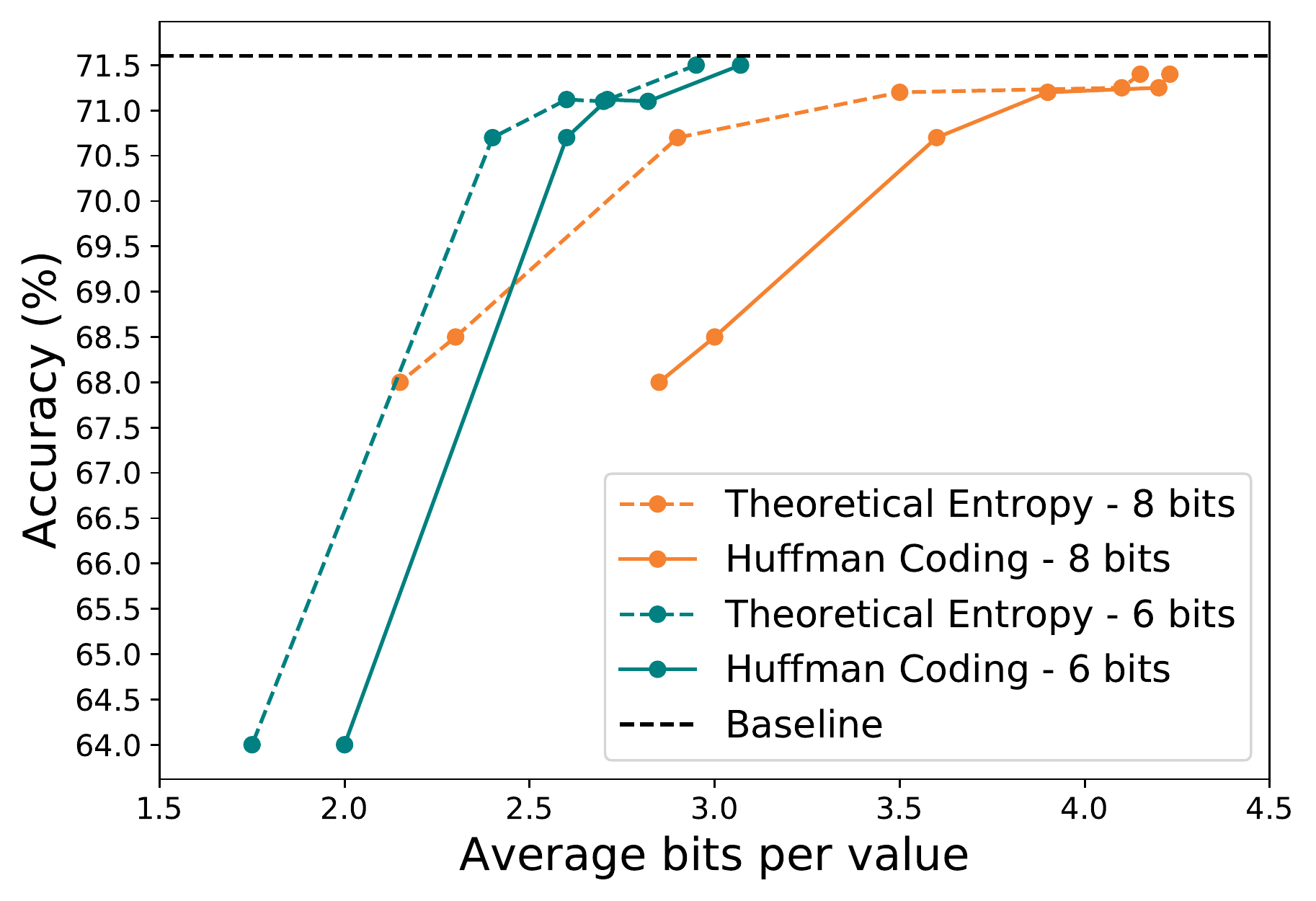}
        \subcaption{ }
        \label{mobilenet-tradeoff}
    \end{subfigure}
 \caption{Tradeoff between rate and accuracy for different values of $\lambda$ (ranged between 0 and 0.3) in \textbf{(a) ResNet-18}  and \textbf{(b) MobileNetV2}. In ResNet-18 the activations are quantized to 5 bits, in MobileNet we show results for activation quantized to 6 and 8 bits. }
\label{fig:acc_entropy}\vspace{-4mm}
\end{figure}

\paragraph{Rate-Accuracy Tradeoff} The proposed CAT algorithm tries to balance between minimizing the rate and maximizing the accuracy of the network by means of the parameter $\lambda$ in \cref{eq:losses}. To check this tradeoff, we run the same experiment in ResNet-18 and MobileNetV2 with different values of $\lambda$ in the range of $0-0.3$, with results shown in \cref{fig:acc_entropy}. 
We show the results of the theoretical entropy and of the Huffman coding, which was chosen for its simplicity but more efficient encoders can be combined with our method. For higher rate Huffman coding is close ($\sim$3\% overhead) to the theoretical entropy, while for lower entropy the difference is higher and is bounded by 1 bit per value -- in the latter case, different lossless coding schemes such as arithmetic coding can provide better results.

\paragraph{Robustness}
For checking the robustness of our method, we performed several runs with the same hyper-parameters and a different random seed. Statistics reported in \cref{tab:robres} suggest that the method is robust and stable under random intialization.
%\ez{in particular, starting from same checkpoint contributes to stability}
\begin{table}
    \centering
    \caption{Mean and standard deviation over multiple runs of ResNet-18 and ResNet-34.}
    \begin{tabular}{lcccc}
        \toprule
        \textbf{Architecture} &\textbf{Compute, bits} & \textbf{Runs}&	\textbf{Accuracy, \%}&	\textbf{Memory, bits} \\
        && &	\textbf{(mean$\pm$std)}&	\textbf{(mean$\pm$std)} \\
            \midrule
            ResNet-18 &5& 5&	69.122$\pm$0.016 &1.5150$\pm$0.0087\\
            ResNet-34 &5&4&73.025$\pm$0.095&1.7875$\pm$0.033\\
            \bottomrule            
    \end{tabular}
    \label{tab:robres}
\end{table}
% \paragraph{Annealing}
% Following the common approach \ez{reference}, we attempted to gradually increase the temperature of soft approximation, applying exponential schedule. However, it didn't affect the results.
% \BCH{clarify}
\paragraph{Soft entropy vs.\ compressibility loss}
Replacing the soft entropy with a different loss which minimizes the entropy almost did not affect the results, as shown in \cref{tab:comp_ent}. We conclude that the desired effect is a result of entropy reduction rather than a particular form of regularization promoting it.
\begin{table}
    \centering
    \caption{Performance of  soft entropy (\ref{eq:diff_entropy})  and compressibility loss (\ref{eq:comp_loss}) on ResNet-18.}
    \begin{tabular}{cccc}
        \toprule
        \textbf{Compute, bits}& \textbf{Loss} &	\textbf{Accuracy}&	\textbf{Memory, bits} \\
            \midrule
            4&entropy&67.86\%&1.43\\
            4&compressability&67.84\%&1.50\\
            \midrule
            5&entropy&69.49\%&1.79\\
            5&compressability&69.36\%&1.73\\
            \bottomrule            
    \end{tabular}
    \label{tab:comp_ent}
\end{table}
\paragraph{Batch size}
We have noticed that training ResNet-50 on a single GPU mandated the use of small batches, leading to performance degradation. Increasing the batch size from 16 to 64 without changing other hyperparameters increased accuracy by more than 0.5\% with entropy increase by less than 0.1 bits/value.

% \ez{todo}.
% \begin{table}
%     \centering
%     \caption{Performance of  different batch sizes on ResNet-50. \ez{merge with \cref{tab:comp_ent}?}}
%     \begin{tabular}{cccc}
%         \toprule
%         \textbf{A BW}& \textbf{Batch size} &	\textbf{Accuracy}&	\textbf{BPP} \\
%             \midrule
%             4&16&73.80\%&1.72\\
%             4&64&74.50\%&1.78\\
%             \midrule
%             5&16&75.20\%&1.70\\
%             5&64&XX\%&\\
%             \bottomrule            
%     \end{tabular}
%     \label{tab:batch_size}
% \end{table}

%\input{sections/05_hardware.tex}
\section{Discussion}
\label{sec:disc}
%Deep networks are increasingly used for applications at the edge. Devices at the edge have many restrictions on processing, memory, power-consumption, and storage for models. 
Quantization of activations reduces memory access costs which are responsible for a significant part of energy footprint in NN accelerators.
%for reading and storing intermediate activations. 
Conservative quantization approaches, known as post-training quantization, take a model trained for full precision and directly quantize it to 8-bit precision. These methods are simple to use and allow for quantization with limited data. Unfortunately, post-training quantization below 8 bits usually incurs significant accuracy degradation. Quantization-aware training approaches involve some sort of training either from scratch \cite{hubara2016quantized}, or as a fine-tuning step from a pre-trained floating point model \cite{han2015deep}.  Training usually compensates significantly for model's accuracy loss due to quantization. 
%Yet, these approaches are usually limited to 4-bit precision. 

In this work, we take a further step and propose a compression-aware training method to aggressively compress activations to as low as $2$ average bit/value representations without harming accuracy. Our method optimizes the average-bit per value needed to represent activation values by minimizing the entropy. We demonstrate the applicability of our approach on classification tasks using the models  MobileNetV2, ResNet, and object detection task using the model SSD512. 
%The proposed method is implemented in hardware, demonstrating \ez{XX} energy consumption reduction. 
Due to the low overhead, the method provides universal improvement for any custom hardware, being especially useful for the accelerators with efficient computations, where memory transfers are a significant part of the energy budget. We show that the effect is universal among loss functions and robust to random initialization.

% We propose training method of reducing bandwidth which minimizes costs at inference time and reduces memory bandwidth ~ twice vs no training and 3+ times for no compression. 
% We show results on MobileNetV2, ResNet, SSD512. show low entropy without performance loss. we implemented the proposed method in hardware, demonstrating \ez{XX} energy consumption reduction. Due to low overhead, the method provides universal improvement for any custom hardware, being especially useful for the accelerators with efficient computations, where memory transfers are a significant part of the energy budget.
% We show that the effect is universal among loss functions and robust to random initialization.

%\subsubsection*{Author Contributions}
%\subsubsection*{Acknowledgments}
%The research was funded by ERC StG RAPID and Hiroshi Fujiwara Technion Cyber Security Research Center.

\vskip 0.2in
\clearpage

\bibliography{training_bandwidth}
\bibliographystyle{iclr2020_conference}

\newpage
\appendix
\crefalias{section}{appsec}
\crefalias{subsection}{appsec}
\crefalias{subsubsection}{appsec}

%%% FIGURE NUMBERING IN APPENDIX
\renewcommand\thefigure{\thesection.\arabic{figure}} 
\renewcommand\thetable{\thesection.\arabic{table}} 
\renewcommand\theequation{\thesection.\arabic{equation}}  
\setcounter{figure}{0}  
\setcounter{table}{0}

\section{Additional results}
\label{app:results}
We list results of the additional experiments we performed in \cref{tab:allres_rn,tab:allres_mn}.

\begin{table}[hb]
    \centering
    \caption{Experimental results for ResNet.}
    %\begin{adjustbox}{max height=0.4\textheight}
    \begin{tabular}{lcccccc}
        \toprule
        \textbf{Architecture} & \textbf{Batch size}&	\textbf{lr}&	\textbf{$\lambda$}&\textbf{Compute}  & \textbf{Memory} &  \textbf{Accuracy, \%} \\
            \midrule
           \multirow{19}{*}{ResNet-18} & \multirow{19}{*}{96} & \multirow{19}{*}{0.001} & 0    & \multirow{3}{*}{4} & 2.050 & 68.000 \\
                                       &    &       & 0.05 &   & 1.540 & 67.950 \\
                                       &    &       & 0.08 &   & 1.430 & 67.860 \\ \cmidrule{4-7}
                                       &    &       & 0.05 & \multirow{9}{*}{5} & 1.790 & 69.490 \\
                                       &    &       & 0.05 &   & 1.750 & 69.400 \\
                                       &    &       & 0.1  &   & 1.410 & 69.120 \\
                                       &    &       & 0.12 &   & 1.361 & 68.900 \\
                                       &    &       & 0.15 &   & 1.280 & 68.914 \\
                                       &    &       & 0.18 &   & 1.120 & 68.160 \\
                                       &    &       & 0.2  &   & 1.110 & 68.300 \\
                                       &    &       & 0.25 &   & 1.040 & 67.800 \\
                                       &    &       & 0.3  &   & 0.974 & 67.700 \\  \cmidrule{4-7}
                                       &    &       & 0    & \multirow{3}{*}{6} & 3.100 & 70.000 \\
                                       &    &       & 0.05 &   & 1.930 & 69.710 \\
                                       &    &       & 0.08 &   & 1.700 & 69.500 \\ \cmidrule{4-7}
                                       &    &       & 0.05 & 7 & 2.280 & 69.660 \\ \cmidrule{4-7}
                                       &    &       & 0    & \multirow{3}{*}{8} & 5.100 & 69.900 \\
                                       &    &       & 0.05 &   & 2.460 & 69.820 \\
                                       &    &       & 0.08 &   & 2.410 & 69.110 \\
            \midrule
            
            \multirow{3}{*}{ResNet-34} & \multirow{3}{*}{96} & \multirow{3}{*}{0.001} & 0.05 & 8 & 2.750 & 73.200 \\
			          &  &  & 0.05 & 6 & 2.000 & 73.200 \\
			          &  &  & 0.05 & 5 & 1.790 & 73.100 \\
            \midrule
            \multirow{8}{*}{ResNet-50}  & 16 & \multirow{8}{*}{0.0001} & 0    & \multirow{4}{*}{4} & 2.500 & 73.700 \\
			            & 16 &        & 0.05 &   & 1.720 & 73.800 \\
			            & 64 &        & 0.05 &   & 1.78  & 74.5   \\
			            & 48 &        & 0.08 &   & 1.67  & 74.2   \\ \cmidrule{4-7}
			            & 16 &        & 0    & \multirow{4}{*}{5} & 2.950 & 75.500 \\
			            & 16 &        & 0.05 &   & 1.920 & 75.460 \\
			            & 16 &        & 0.08 &   & 1.700 & 75.200 \\
			            & 16 &        & 0.1  &   & 1.600 & 74.900 \\
            \bottomrule            
    \end{tabular}
%\end{adjustbox}
    \label{tab:allres_rn}
\end{table}

\begin{table}[hb]
	\centering
	\caption{Experimental results for MobileNet.}
	%\begin{adjustbox}{max height=0.4\textheight}
	\begin{tabular}{lcccccc}
		\toprule
		\textbf{Architecture} & \textbf{Batch size}&	\textbf{lr}&	\textbf{$\lambda$}&\textbf{Compute}  & \textbf{Memory} &  \textbf{Accuracy, \%} \\
		\midrule
		\multirow{24}{*}{MobileNetV2} & 64 & 0.0001 & 0    & \multirow{6}{*}{4} & 2.200 & 66.150 \\
		& 64 & 0.001  & 0    &   & 2.800 & 66.200 \\
		& 64 & 0.0001 & 0.05 &   & 2.100 & 66.900 \\
		& 64 & 0.001  & 0.05 &   & 2.080 & 66.400 \\
		& 64 & 0.001  & 0.08 &   & 1.830 & 66.200 \\
		& 64 & 0.0001 & 0.08 &   & 1.980 & 66.450 \\ \cmidrule{4-7}
		& 64 & 0.001  & 0    & \multirow{8}{*}{6} & 3.900 & 69.600 \\
		& 64 & 0.0001 & 0    &   & 3.700 & 71.000 \\
		& 96 & 0.0001 & 0.05 &   & 2.950 & 71.500 \\
		& 32 & 0.0001 & 0.08 &   & 2.700 & 70.930 \\
		& 64 & 0.0001 & 0.1  &   & 2.700 & 70.950 \\
		& 32 & 0.001  & 0.15 &   & 1.750 & 64.000 \\
		& 64 & 0.0001 & 0.15 &   & 2.600 & 71.200 \\
		& 64 & 0.0001 & 0.2  &   & 2.450 & 70.700 \\ \cmidrule{4-7}
		& 64 & 0.0001 & 0    & \multirow{10}{*}{8} & 4.750 & 71.300 \\
		& 64 & 0.0001 & 0.05 &   & 4.150 & 71.400 \\
		& 64 & 0.001  & 0.08 &   & 2.600 & 70.000 \\
		& 64 & 0.0001 & 0.08 &   & 4.120 & 71.600 \\
		& 64 & 0.001  & 0.1  &   & 2.400 & 69.600 \\
		& 64 & 0.0001 & 0.1  &   & 4.100 & 71.250 \\
		& 64 & 0.001  & 0.15 &   & 2.300 & 68.500 \\
		& 64 & 0.001  & 0.2  &   & 2.150 & 68.000 \\
		& 64 & 0.0001 & 0.2  &   & 3.500 & 71.200 \\
		& 32 & 0.0001 & 0.3  &   & 2.900 & 70.700 \\
		\bottomrule            
	\end{tabular}
	%\end{adjustbox}
	\label{tab:allres_mn}
\end{table}

\end{document}